\documentclass[pmlr,twocolumn,10pt]{jmlr}

\mlhtrack{proceedings}

\newif\iffinal
\finalfalse  

\iffinal
    \ifmlhneedspmlr
      \jmlrvolume{XXX}
      \jmlryear{2026}
    \fi
    \ifmlhfindings \jmlrproceedings{}{ML4H 2026 - Findings Track}\fi
    \ifmlhdemo     \jmlrproceedings{}{ML4H 2026 - Demo Track}\fi
    \jmlrworkshop{Machine Learning for Health (ML4H) 2026}
\else
    \jmlrproceedings{}{Preprint}
    \jmlrworkshop{}
\fi

\usepackage{booktabs}
\usepackage{siunitx}
\usepackage[switch]{lineno}
\usepackage{tikz}
\usetikzlibrary{arrows.meta,positioning,calc}
\usepackage{multirow}

\newcommand{\model}{Sentry}  
\newcommand{\rc}{\textsc{RC}}
\newcommand{\adapt}{\textsc{Adaptive}}
\newcommand{\pOnly}{\textsc{P-only}}
\DeclareMathOperator{\CP}{CP}

\title[Certified AI Triage of ICU Alarms]{Certified AI Triage of ICU Alarms}

\author{%
  \Name{Mohammed Sameer Syed} \Email{mohammed@roshan-ai.com}\\
  \addr Roshan AI
  \AND
  \Name{Rozhin Yasaei} \Email{yasaei@arizona.edu}\\
  \addr University of Arizona
}

\begin{document}

\maketitle

\ifmlhdemo\else
\begin{abstract}
In the VTaC benchmark 71\% of ventricular-tachycardia alarms are false, but silencing a real one can delay recognition of a dangerous arrhythmia. We reframe alarm reduction as three-way triage (retain, suppress, or defer) and bound the decision this analysis treats as harmful: among suppressed alarms, the fraction that were genuine stays below a user-set budget with 95\% confidence, under i.i.d.\ event sampling. Alarms sharing a waveform record are dependent, so the clustered analysis is a sensitivity check. On the official split a 5\% budget certifies in all three seeds, suppressing 74.8\% of false alarms while silencing 1.5\% of genuine ones, at AUROC 0.953 and Challenge Score 83.33, numerically comparable to the strongest of the eleven published systems. Our central finding measures what multiplicity costs: the correction charges for every candidate, so a finer grid can certify strictly less. Under held-out calibration the 885-cell grid we declared certifies 1 of 15 fold-runs, while choosing the grid on a separate selection partition certifies 8. We project the calibration volume each budget needs, making an uncertifiable budget a design parameter. Finally, adding a learned reliability dimension to the policy grid did not sharpen the certified frontier.
\end{abstract}
\begin{keywords}
  false alarm reduction, ventricular tachycardia, selective prediction, risk control, multimodal physiological waveforms, VTaC
  \end{keywords}
  \fi
  
  \ifmlhneedsstatements
  \paragraph*{Data and Code Availability}
  All experiments use the public VTaC v1.0 release on PhysioNet \citep{lehman2024vtacphysionet,lehman2023vtac}, open access under CC~BY-SA~4.0; no other data were scored. Code is not included as supplementary material, will be fully released upon acceptance.
  
  \paragraph*{Institutional Review Board (IRB)}
  A retrospective analysis of a de-identified public benchmark, this work did not require IRB approval. The software is a research artifact and is not connected to any clinical system.
  \fi

\section{Introduction}
\label{sec:intro}

Bedside monitors in intensive care units generate arrhythmia alarms faster than clinical attention can absorb, and most are false \citep{drew2014insights,sendelbach2013alarm,cvach2012monitor}. Ventricular tachycardia (VT) is the consequential case: it can be life-threatening, so silencing a true VT alarm can delay recognition of a dangerous arrhythmia, and VT has been among the harder alarm types to adjudicate automatically \citep{clifford2016false,lehman2023vtac}. The VTaC benchmark \citep{lehman2023vtac} provides 5{,}037 expert-adjudicated VT alarms with multi-lead ECG and pulsatile waveforms.

A classifier threshold trades sensitivity against specificity but never states when the evidence is too incomplete or degraded to act on. In retrospective adjudication the natural output is three-way (\emph{retain}, \emph{suppress}, or \emph{defer} for review); unsafe suppression is the endpoint this analysis controls. What a deployment needs is a finite-sample bound on that rate and an account of what the bound costs in labelled data. This paper makes five contributions on VTaC.
\begin{enumerate}
\item \textbf{A safety guarantee for VT-alarm suppression.} We frame adjudication as choosing a retain/suppress/defer policy from a predeclared grid and certify it with Learn-then-Test \citep{angelopoulos2025learn} under i.i.d.\ event sampling, with a record-clustered bootstrap probing the dependence that assumption ignores. To our knowledge no published VTaC system reports a finite-sample bound on the harmful error.

\item \textbf{What multiplicity costs on a clinical policy grid.} That Bonferroni charges per candidate is known \citep{zecchin2025altt}; the size of the bill on a real task is not. A finer grid searches more policies but pays a stricter level for each, so past some point more candidates certify less: choosing the grid on the selection partition rather than by convention takes certified fold-runs from 1 of 15 to 8 at $\alpha=0.05$, and replicates on the official split.

\item \textbf{Certification without a discrimination penalty.} The base scores are untouched by the policy layer, and on the official split they are comparable to the strongest published baseline on every metric we report.

\item \textbf{Certification cost made computable.} When a budget cannot be certified we report whether the sample or the model appears to be the obstacle, and project how many labelled alarms it would need.

\item \textbf{A negative result the framework can express.} A learned evidence-reliability score $r$ responds to degradation that barely moves the class probability, yet gating suppression on $r$ improved the certified frontier on neither clean nor degraded cohorts, nor anywhere on the grid surface.

\end{enumerate}
Every experimental number here is regenerated from one stored artifact by code holding no result literals; \appendixref{apd:corrections} records an earlier audit and the corrections it forced.

\section{Related Work}
\label{sec:related}

\paragraph{False alarm reduction.} The PhysioNet/CinC 2015 challenge \citep{clifford2015physionet,clifford2016false} established multimodal analysis for false arrhythmia alarms, its strongest entries corroborating ECG beats with pulsatile waveforms \citep{plesinger2016taming,aboukhalil2008reducing} and weighting channels by signal quality \citep{li2008signal}. VTaC \citep{lehman2023vtac} is an order of magnitude larger, independently labelled by at least two experts with adjudication, and reports supervised, contrastive, and generative baselines on a fixed record-level split; we adopt its real-time input and scoring protocol, with the deviations named in \appendixref{apd:protocol}. A complementary line pretrains quality-aware waveform encoders: QualityFM \citep{guo2025qualityfm} self-distills across paired high- and low-quality signals and reports VT false-alarm detection among its transfer tasks. We do not compare encoders; a stronger backbone would help where certification is model-limited, but \tableref{tab:calreq} finds almost every failure here to be sample-limited instead.

\paragraph{Selective prediction and risk control.} Abstention with a reject option dates to \citet{chow1970optimum}; \citet{geifman2017selective} bound selective risk for a given confidence-rate function, already correcting a confidence budget across the thresholds their search visits \citep[see also][]{elyaniv2010foundations}. What differs here is the two-dimensional family, the action-specific endpoint, and the size of grid that correction must cover. Conformal and risk-controlling methods \citep{vovk2005algorithmic,bates2021distribution,angelopoulos2024conformal,angelopoulos2023gentle} supply such guarantees, and Learn-then-Test \citep{angelopoulos2025learn} extends them to non-monotone losses by treating each candidate as a hypothesis test. We apply it to a two-dimensional family and, unlike the common fixed-sequence variant, correct over the full grid (\sectionref{sec:certification}). That Bonferroni charges for every candidate is known, and adaptive Learn-then-Test \citep{zecchin2025altt} answers it with e-process sequential testing \citep[surveyed by][]{farzaneh2026hyperparameter}. Its saving is in testing \emph{rounds}, each buying new data; under a non-adaptive policy evaluated at the horizon it coincides with Learn-then-Test, which is our regime, one frozen sample scoring all 885 candidates at once. We therefore measure the cost rather than remove it. Closest to the quantity bounded here is selective conformal risk control \citep{xu2025selective}, which bounds a prediction-set loss conditional on acceptance where we bound the rate of one harmful action among the events acted on; its search over first-stage thresholds pays the same multiplicity \sectionref{sec:grid} measures.

\paragraph{Missing modalities.} Modality dropout \citep{neverova2016moddrop} improves robustness to absent channels, and \citet{ma2022multimodal} show transformers degrade sharply under missing modalities unless fusion is designed for it; 6\% of VTaC events have no usable pulsatile channel, so every model here carries a mask.

\section{Data and Evaluation Protocol}
\label{sec:data}

\paragraph{Dataset.} VTaC v1.0 \citep{lehman2024vtacphysionet} contains 5{,}037 VT alarm events from ICU monitors of three manufacturers in three US hospitals at 250\,Hz, five minutes before and one after alarm onset. Modality coverage is not uniform: every event carries at least one ECG lead, but some lack a second and some carry no pulsatile channel (PLETH or ABP), so the benchmark is variable-modality by construction rather than by accident.\footnote{The v1.0 release contains 90{,}000 samples per event at 250\,Hz, i.e.\ six minutes with onset at 300\,s, where the benchmark paper's text describes ten; all windows here are measured backwards from onset, so no input is affected.} Each event has a final adjudicated label $y\in\{0,1\}$, with $y{=}1$ denoting a true alarm: 1{,}441 true alarms (28.6\%) and 3{,}596 false. Events belong to 2{,}260 waveform records (2.23 events per record). The release exposes no patient identifier, so record-disjoint separation is the strongest achievable; we describe it as \emph{record-safe}, never patient-safe.

\paragraph{Two arms.} Both arms share architecture, objectives, optimizer, and policy family, differing only in the input contract the published benchmark fixes and the one architectural consequence it forces, the token stride (\sectionref{sec:model}). The \textbf{development arm} pools all 5{,}037 events into five record-disjoint outer folds, each split record-disjointly into training, checkpoint-selection, policy-selection and policy-certification roles with the outer fold as evaluation, on a 60\,s pre-alarm window in three slots (two ECG, one pulsatile); three seeds (317, 911, 2718) give 15 fold-runs. Because the official test partition is dissolved into the pool, \emph{no number from this arm is comparable to a published VTaC result}. The \textbf{official-split arm} honours the published partition exactly (test: 482 events across 226 records, 137 true) and follows the published real-time protocol \citep{lehman2023vtac}: a 10\,s window, four channels with zero imputation, per-segment z-normalization, and the Challenge Score $(\mathrm{TP}+\mathrm{TN})/(\mathrm{TP}+\mathrm{TN}+\mathrm{FP}+5\,\mathrm{FN})$ at the validation-maximizing threshold, with the deviations of \appendixref{apd:protocol}. Its calibration partitions (819 + 842 events) are cut from official \emph{train} only, so the test set is never used for selection or certification. Split assignments are committed, hash-verified artifacts checked for disjointness at every stage.

\paragraph{Preprocessing.} ECG channels receive a 60\,Hz notch and a 1--30\,Hz Butterworth band-pass, PLETH a 0.5--5\,Hz band-pass, ABP none; each is z-scored per event. A slot is \emph{available} if a matching header channel exists and holds at least one finite sample in the window; unavailable slots are zero-filled and flagged in the mask $\mathbf{m}\in\{0,1\}^{C}$. A slot named in the header can still fail this test, so the development cache has 45 events with a single usable ECG slot and 304 with no usable pulsatile slot, against 44 and 292 by header alone. All remain in the evaluation population.

\section{Method}
\label{sec:method}

\begin{figure*}[t]
\floatconts
  {fig:arch}
  {\caption{The \model{} inference path, official-split arm ($C{=}4$, $T{=}312$, $d{=}512$; 14.89\,M parameters); training objectives are in \appendixref{apd:arch}. Panel 4 is the fusion mask: attention spans all $4T$ tokens, each slot pair within one temporal band, so a pulsatile token reaches nearby ECG tokens but never the whole window. The drawn episode lacks a usable pulsatile slot, zeroed by $\mathbf{m}$ at panel 3; 65\% of episodes lack at least one. Traces are schematic; panel 6's segment widths are not to scale.}}
  {\includegraphics[width=0.62\textwidth]{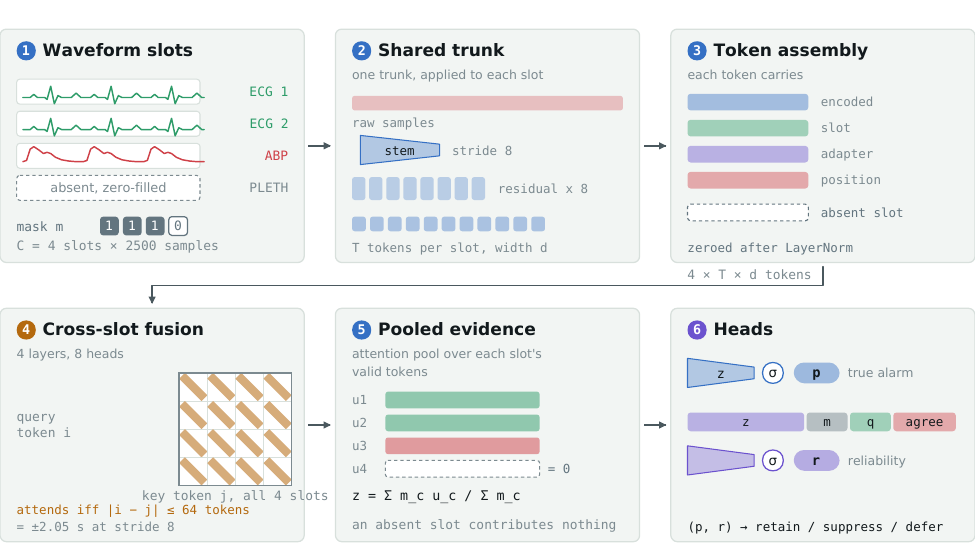}}
\end{figure*}

\subsection{\model{}: availability-aware multimodal encoder}
\label{sec:model}

\figureref{fig:arch} summarizes the model. Each event provides waveforms $\mathbf{x}\in\mathbb{R}^{C\times L}$ over $C$ modality slots of $L$ samples, and an availability mask $\mathbf{m}$. A trunk shared across slots produces $T$ tokens per slot at width $d=512$; tokens of unavailable slots are zeroed throughout, and attention pooling gives one vector per slot, unavailable slots pooling to zero rather than to a value derived from imputed zeros. Fusion is four self-attention layers over the concatenated tokens, masked so a pulsatile token can attend to nearby ECG tokens but never globally over raw samples. The fused evidence vector $z\in\mathbb{R}^{d}$ is the availability-weighted mean of the slot-pooled outputs, and a classifier head maps it to the true-alarm probability $p=\sigma(\cdot)$, with $\sigma$ the logistic function. The token stride is $s=40$ in the development arm and $s=8$ in the official-split arm: the shorter window forces the change, since at $s=40$ a 10\,s window yields 62 tokens, fewer than the $\pm64$-token fusion radius, so local attention would degenerate to global.

The \textbf{reliability head} consumes $z$, the mask $\mathbf{m}$, per-slot reliability scores $q\in\mathbb{R}^{C}$ from a channel head, and a shift-compatibility confidence: the maximum softmax over cosine agreement between the first ECG slot and each other slot across 17 token shifts ($\pm8$ tokens, $\pm1.3$\,s at stride 40 and $\pm0.26$\,s at stride 8). Its $d+3C-1$ inputs (520 development, 523 official) are dominated by $z$. Compatibility is a learned agreement feature, not a pulse-transit-time estimate, and $r=\sigma(\cdot)$ describes the evidence, not physiology.

\paragraph{Objectives.} Training minimizes unweighted binary cross-entropy under modality dropout, plus three auxiliary losses on a separately intervened view of the same batch: masked cross-modal reconstruction for dropped slots, per-slot identification of a declared acquisition intervention, and an \emph{evidence-quality} target, the fraction of source-available slots left unmasked and unintervened in that view, regressed by the reliability logit. Every target is an observable consequence of an intervention the training loop applied; none uses the VT label. \appendixref{apd:arch} gives the trunk, loss weights and optimizer settings.

\subsection{Triage policy family}
\label{sec:policy}

A policy is a triple $(\tau_{\mathrm{sup}},\tau_{\mathrm{rel}},\tau_{\mathrm{ret}})$. An event is \emph{suppressed} if $p\le\tau_{\mathrm{sup}}$ and $r\ge\tau_{\mathrm{rel}}$, \emph{retained} if $p\ge\tau_{\mathrm{ret}}$ (retain wins any overlap), and \emph{deferred} otherwise. The harmful event is a true alarm assigned \emph{suppress}; the \textbf{unsafe-suppression risk} is $\Pr(y=1\mid \text{suppress})$. Retention takes precedence, so the suppressed set is $\{p\le\tau_{\mathrm{sup}},\,r\ge\tau_{\mathrm{rel}},\,p<\tau_{\mathrm{ret}}\}$, independent of $\tau_{\mathrm{ret}}$ exactly when $\tau_{\mathrm{ret}}>\tau_{\mathrm{sup}}$ makes the overlap empty. It holds in all 172 policies selected across the 216 pooled arm-by-budget-by-criterion-by-family cells, so the certified set is the deployed set there; we verify this rather than assume it. Split runs record no thresholds, so we do not claim it for them. The retain threshold is fixed afterwards as the largest value on a 91-point grid retaining at least 95\% of true alarms on the policy-selection partition. The predeclared suppression grid crosses 59 probability thresholds (20 in $[0.0005,0.01]$ and 40 in $[0.01,0.40]$, the shared 0.01 endpoint counted once) with 15 reliability thresholds in $[0.20,0.90]$: 885 candidates. We call this full two-dimensional family \rc{}, for reliability-conditioned. The \emph{probability-only comparator} (\pOnly{}) pins $\tau_{\mathrm{rel}}$ to 0 and is otherwise identical, isolating the contribution of $r$; its 59 candidates receive a correspondingly milder correction.

\subsection{Certification by Learn-then-Test}
\label{sec:certification}

For budget $\alpha$ and confidence $1-\delta=0.95$, each candidate $\lambda$ in the grid $\Lambda$ is tested on the calibration sample: with $n_\lambda$ suppressions and $k_\lambda$ of them true alarms, the exact one-sided Clopper--Pearson upper bound \citep{clopper1934use} $\CP^{+}(k_\lambda,n_\lambda;\,\delta/|\Lambda|)$ must satisfy $\CP^{+}\le\alpha$. Among admitted candidates the one with most false-alarm suppressions is selected; by Bonferroni \citep{bonferroni1936teoria} over $\Lambda$ its true risk is at most $\alpha$ with probability $\ge 1-\delta$ under i.i.d.\ sampling \citep{angelopoulos2025learn}. If no candidate is admitted the fold is \emph{infeasible} and contributes a null (suppress-nothing) policy to averages. Throughout, a run is \emph{certified} at $\alpha$ when this test admits a candidate: the term refers to the corrected bound being met on the calibration sample under the stated assumptions, not to the risk observed later on evaluation data.

\paragraph{Why not fixed-sequence testing.} Suppression sets are nested in $\tau_{\mathrm{sup}}$, which invites the fixed-sequence variant that walks thresholds in a predeclared order and stops at the first failure, spending no correction at all. It would be valid under any predeclared order, but has no power here. Even if the population conditional risk were non-decreasing in $\tau$, the Clopper--Pearson acceptance region is not, because the bound depends on the suppression-set size $n_\tau$ as well as the unsafe count $k_\tau$: at small $\tau$ the set is tiny and the bound wide for want of data, not because risk is high. The region is an \emph{interval} interior to the ordering rather than a prefix, so an ascending walk halts at its first step in 15 of the 18 runs at $\alpha{=}0.05$, and for a reason that does not depend on the level at which the step is tested: the smallest threshold suppresses \emph{zero} events, which forces a Clopper--Pearson upper bound of 1.0 before risk is consulted at all (\appendixref{apd:fixedseq}). We did not run the descending order at the uncorrected level, so we claim no result for it. Bonferroni assumes no ordering and pays in level rather than reachability.

\paragraph{Calibration mode.} Learn-then-Test derives its guarantee from the sample it selects on, so we pool the policy-selection and policy-certification partitions into one calibration sample (1{,}555 events, development; 1{,}661, official). The ``Cert.'' counts in \tableref{tab:triage} are therefore a feasibility record rather than an independent check. We also run a held-out variant (\texttt{split} calibration), selecting on one partition and bounding on the disjoint other; Learn-then-Test does not require this and splitting halves the calibration data, so that column is the stricter of the two rather than the correct one.

\paragraph{Record clustering.} The Clopper--Pearson step needs the suppressed outcomes to be independent Bernoulli draws. Events from one record share patient, device, and acquisition conditions, and record-disjoint partitions keep a record out of two roles without making the events inside one record independent, so that assumption does not hold exactly on VTaC. We therefore compute a record-clustered bootstrap (2{,}000 record resamples) \citep{field2007bootstrapping} and report it as a \emph{sensitivity analysis}, not a second certificate. Three limits are worth stating. A percentile bootstrap is an approximate uncertainty assessment rather than an exact distribution-free bound; 2{,}000 resamples cannot resolve a tail at the corrected level, since $\delta/885$ leaves 0.11 expected draws in it; and when no suppressed alarm is genuine the resampled risk is identically zero, so its upper percentile is zero however few events were seen. That it changes no admission decision is agreement between two procedures, not evidence of coverage. A second, distinct bootstrap produces the benchmark intervals in \tableref{tab:base}; \appendixref{apd:protocol} defines both. No patient identifier is released, so a patient-level bound is not computable and record independence is itself an assumption.

\paragraph{Sample-limited versus model-limited.} For each budget we compute, from frozen evaluation predictions, the smallest calibration size admitting some grid cell at the corrected level in \emph{every} run. A budget looks \emph{model-limited} if observed risk there already exceeds it, and \emph{sample-limited} if the point estimate meets it but the bound does not.

\section{Results}
\label{sec:results}

All values are read from one generated input artifact built from stored run outputs; fold spreads are descriptive SDs, since folds share one dataset \citep{roberts2017cross}.

\subsection{Base prediction}

\begin{table}[htbp]
\floatconts
  {tab:base}
  {\caption{Base predictor: mean $\pm$ SD over 15 development fold-runs (15{,}111 evaluations, 4{,}323 true), and mean over 3 seeds on the official test set (482 events, 137 true) with 95\% record-clustered intervals; across-seed SD there is 0.005 AUROC, 0.026 AUPRC, 0.25 Score, 0.012 F1. ``Score'' is the Challenge Score $\times100$. A dash marks a quantity a column does not report: Score and F1 need the published validation-selected threshold, which the development arm does not define, and \citet{lehman2023vtac} report no AUPRC. Their ten other real-time systems are in \tableref{tab:basefull}.}}
  {\small
  \setlength{\tabcolsep}{3pt}
  \begin{tabular}{lcccc}
  \toprule
  & Development & Official & 95\% CI & FCN$^{\dagger}$ \\
  \midrule
  AUROC & 0.926\,$\pm$\,0.013 & 0.953 & [0.922, 0.970] & 0.949 \\
  AUPRC & 0.843\,$\pm$\,0.032 & 0.875 & [0.793, 0.925] & -- \\
  Score & -- & 83.33 & [78.5, 86.8] & 80.08 \\
  F1    & -- & 0.799 & [0.729, 0.852] & 0.805 \\
  \bottomrule
  \multicolumn{5}{l}{\scriptsize $^{\dagger}$Transcribed; no per-event predictions.}
  \end{tabular}}
\end{table}

\tableref{tab:base} reports discrimination. The supportable claim on the official split is parity with the best published baseline, not superiority: FCN \citep{wang2017time} lies inside our 95\% record-clustered interval on AUROC, Score, and F1, and no per-event predictions exist for a paired test. Of the ten other published systems (\tableref{tab:basefull}), only CNN+CL's Score sits inside that interval, just above its lower edge; the other nine fall below it. The seed SD is an order of magnitude smaller than the clustered interval and is not uncertainty about the estimate. \appendixref{apd:protocol} adds calibration and operating-point metrics.

\subsection{Certified triage}

\begin{table}[htbp]
\floatconts
  {tab:triage}
  {\caption{Learn-then-Test certification at 95\% confidence, Bonferroni over 885 candidates (\rc{}) or 59 (\pOnly{}). ``Cert.'' is a feasibility count on the pooled sample, not an independent check; ``Held-out'' repeats it under \texttt{split} calibration (\sectionref{sec:certification}). Columns average over all fold-runs, infeasible ones contributing a null policy; the ``FA supp.'' spread is an SD across the three seed means, not across the fifteen fold-runs. The $\alpha=0.02$ rows are omitted: nothing certifies there except one pooled official seed.}}
  {\scriptsize
  \setlength{\tabcolsep}{3pt}
  \begin{tabular}{lcccccc}
  \toprule
  Policy & $\alpha$ & Cert. & \textbf{Held-out} & FA supp. & Risk & TA kept \\
  \midrule
  \multicolumn{7}{l}{\emph{Development arm (15 fold-runs)}}\\
  \rc{}    & 0.05 & 11/15 & 1/15  & 37.7\,$\pm$\,6.8\% & 0.016 & 0.977 \\
  \pOnly{} & 0.05 & 14/15 & 6/15  & 57.3\,$\pm$\,2.9\% & 0.028 & 0.955 \\
  \rc{}    & 0.10 & 15/15 & 12/15 & 87.6\,$\pm$\,0.9\% & 0.068 & 0.841 \\
  \pOnly{} & 0.10 & 15/15 & 12/15 & 88.9\,$\pm$\,0.8\% & 0.072 & 0.828 \\
  \addlinespace
  \multicolumn{7}{l}{\emph{Official split (3 seeds)}}\\
  \rc{}    & 0.05 & 3/3 & 2/3 & 70.4\,$\pm$\,4.1\% & 0.008 & 0.985 \\
  \pOnly{} & 0.05 & 3/3 & 2/3 & 74.8\,$\pm$\,1.3\% & 0.008 & 0.985 \\
  \rc{}    & 0.10 & 3/3 & 3/3 & 87.4\,$\pm$\,0.6\% & 0.037 & 0.915 \\
  \pOnly{} & 0.10 & 3/3 & 3/3 & 87.4\,$\pm$\,0.6\% & 0.038 & 0.912 \\
  \bottomrule
  \end{tabular}}
\end{table}

\tableref{tab:triage} gives the primary result. We take $\alpha=0.05$ as the reference operating point: it is the tightest budget this calibration sample certifies, and the only one whose unconditional cost to true alarms is small. The four \rc{} failures there are infeasible folds, no grid cell admitted rather than a violated bound. A 2\% budget certifies in no development fold-run and in one pooled official seed only, and the record-clustered sensitivity analysis changes no admission.

Two safety quantities must be read separately, and the distinction is why we headline $\alpha=0.05$. The conditional risk $\Pr(y{=}1\mid\text{suppress})$ is what is certified; the unconditional fraction of true alarms suppressed is what a clinician would ask about, and the two diverge sharply. At $\alpha=0.05$ in the development arm 105 of the 4{,}323 pooled true alarms are silenced; at $\alpha=0.10$ it is 690, because that policy suppresses 10{,}138 alarms in all. A 10\% conditional budget therefore carries a much larger unconditional cost even though it certifies everywhere; every claim here should be read at $\alpha=0.05$.

The official split is best stated in counts: on its 482-event test set the \pOnly{} policy suppresses 265, 259, and 256 alarms across the three seeds, of which 2, 3, and 1 were genuine VT out of the 137 present. The worst seed silences three real arrhythmia alarms and 256 of 345 false ones. That arm survives the held-out check better, certifying two of three seeds against 6 of 15 development fold-runs.

\subsection{Reliability gating does not help}
\label{sec:null}

At every certified budget in both arms, \pOnly{} certifies in at least as many folds and suppresses at least as many false alarms as \rc{} (\figureref{fig:frontier}). The 885-candidate family pays a stricter level and is infeasible in four folds at $\alpha=0.05$ against \pOnly{}'s one, but where both certify the selected \rc{} policies are no better either. At $\alpha=0.10$ on the official split every selected $\tau_{\mathrm{sup}}$ sits at the top of the probability grid (0.39--0.40), so that frontier is grid- rather than risk-limited; at $\alpha=0.05$ they lie between 0.01 and 0.08.

The obvious rebuttal is that clean VTaC is the wrong place to look, since a reliability score can only help where evidence quality varies. On three cohorts in which half the \emph{records} carry a declared degradation (pulsatile dropout, ECG burst artifact, baseline wander), \pOnly{} matches or exceeds \rc{} in 17 of the 18 cohort-by-budget-by-mode cells. Calibrating on clean data and evaluating on a degraded cohort, \rc{} does attain lower realised risk (0.016 against 0.035 under burst artifact at $\alpha=0.05$), but at correspondingly lower suppression: it is more conservative, not better.

The head is not inert, which is what makes this a null result rather than a bug: baseline wander lowers $r$ by 0.317 while moving $p$ by $+0.003$, and mean $r$ separates availability strata that AUPRC does not (\appendixref{apd:stress}). Discrimination and reliability are different quantities, which is why $r$ is a plausible deferral signal and also why, where $p$ is already well separated, gating on it did not improve suppression in any comparison we ran (\appendixref{apd:descriptive}).

\subsection{What Learn-then-Test buys}
\label{sec:baselines}

Is the correction unnecessary machinery? \tableref{tab:proceduresfull} says not. Selecting the largest threshold whose \emph{observed} calibration risk meets the budget admits a policy almost always and then exceeds that budget on held-out data in two thirds of runs at $\alpha=0.05$; adding an exact bound but no multiplicity correction still overshoots in three of fourteen admitted runs. Split conformal never overshoots at $\alpha\ge0.05$ but controls a different functional, the marginal fraction of genuine alarms falling in the suppression region rather than the conditional risk among suppressed ones. Learn-then-Test is the only procedure whose realised risk stayed inside the advertised budget in every admitted run here, though these are observed test proportions and no calibration-time bound promises them. On the official split, with a stronger model and a larger calibration sample, all four stay inside the budget at $\alpha\ge0.02$.

Nor is the multiplicity price an artifact of Bonferroni's crudeness. Holm tests its first hypothesis against $\delta/|\Lambda|$, exactly the Bonferroni threshold, so the two have identical any-rejection feasibility; across 432 run-cells Holm changes no verdict and gains at most 0.87 percentage points of suppression (\appendixref{apd:holm}).

\subsection{The grid is an expensive hyper-parameter}
\label{sec:grid}

\begin{figure}[t]
\floatconts
  {fig:grid}
  {\caption{Held-out certification against grid size, development arm, $\alpha=0.05$. Certified runs are of 15. The one-dimensional grids are one family at increasing resolution and are joined; the two-dimensional grids are distinct combinations of $n_p$ probability levels by $n_r$ reliability levels that merely sort by size, so they are left unjoined. The dashed line is the adaptive rule of \sectionref{sec:adaptive}, annotated with its value.}}
  {\includegraphics[width=\linewidth]{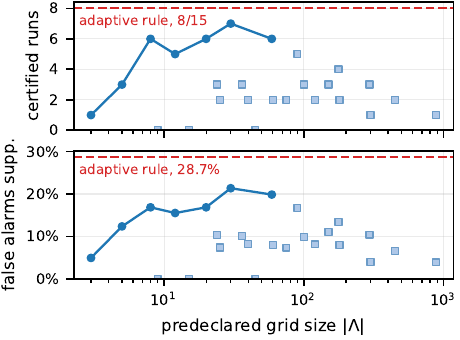}}
\end{figure}

The \rc{} results above use one 885-cell grid, declared once and never revisited, and \figureref{fig:grid} shows what that costs: under held-out calibration at $\alpha=0.05$ that grid certifies 1 of 15 fold-runs at 3.9\% suppression while a 30-cell grid certifies 7 at 21.4\%. The correction is charged per candidate, so a grid fine enough to contain a better policy can be too fine to certify any policy at all. \tableref{tab:grid} gives the ladder and \appendixref{apd:gridsurface} the full surface.

Two structural facts stand out. At matched candidate counts the one-dimensional grids dominate: $8\times1$ certifies 6 runs while $3\times3$ certifies none, and $30\times1$ beats $12\times3$, $59\times1$ beats $12\times5$. Spending candidates on the reliability axis buys nothing and costs multiplicity, as \sectionref{sec:null} finds independently. And no grid on the ladder certifies $\alpha\le0.02$, including the 3-candidate grid whose correction is nearly free: for this model and ladder, coarsening cannot rescue the tightest budgets, and \tableref{tab:calreq} projects what volume might.

\paragraph{The effect replicates on the official split.} The same ladder on the official-split arm, a different model, pipeline, window and calibration sample, reproduces both facts: the 885-cell grid certifies fewer seeds than a 12-cell grid, non-monotonically, and no two-dimensional grid beats the best one-dimensional grid of matched size (\tableref{tab:gridofficial}). The effect is confined to the marginal band: at $\alpha=0.10$ every grid certifies every seed and at $\alpha\le0.02$ none does, so grid size matters only where the budget is close to what the sample supports.

\subsubsection{Choosing the grid on the selection partition}
\label{sec:adaptive}

Reading a winning row off \tableref{tab:grid} would select a hyper-parameter on the evaluation set, so we fix a rule, \adapt{}: score every grid by the false-alarm suppression its best \emph{admissible} candidate reaches on the selection partition, keep the highest-scoring one, smaller on ties, then select and certify with it as before. This spends no data selection was not already using, and the bound stays valid: an uncorrected test of one frozen policy on data the rule never read (\appendixref{apd:leak}). It certifies 8 of 15 development runs against 1 of 15 for the fixed 885-cell grid, silencing 1.4\% of genuine alarms, and every official seed where that grid certifies two. The chosen grids are one-dimensional throughout, from 3 to 59 candidates; in the 7 runs where nothing certifies the rule falls back to the smallest. Its margin over the best fixed one-dimensional grid is one fold (8/15 against 7/15), so its value is not that it beats a well-chosen grid but that it finds one without being told which to declare.

\paragraph{What certification costs.}

\tableref{tab:calreq} projects the pattern from the frozen predictions: with 885 candidates, $\alpha=0.10$ needs about 800 calibration events and $\alpha=0.05$ about 4{,}000 before every run certifies, against the 1{,}555 available. Development-arm failures at $\alpha=0.02$ look sample-limited, the point-estimate risk being below 2\% while no binomial bound on 1{,}555 events certifies it against 885 hypotheses. These are plug-in projections conditional on the observed score distribution, not guarantees that a future sample of that size will certify; a grid should be no finer than the calibration sample can support.

\subsection{One subgroup pays for the marginal guarantee}
\label{sec:conditional}

The guarantee is marginal, and promises nothing about any subpopulation. VTaC releases no manufacturer, site, or patient identifier, so the only subgroup axis it supports is which waveform channels were present; we stratify the evaluation partition by that mask, holding the certified policy fixed.

\tableref{tab:conditional} reports the result. The stratum without a usable pulsatile channel is 6\% of evaluations, and in it the realised risk runs 2.6 times the modality-complete risk at $\alpha{=}0.05$ (16/213 against 180/6{,}133) and 1.7 times at $\alpha{=}0.10$ (53/453 against 684/9{,}777). We report counts: the same events recur across seeds, so a test treating them as independent would overstate the evidence. The budget is exceeded there in two of three seeds at $\alpha{=}0.05$ and in all three at $\alpha{=}0.10$, while the marginal risk meets it every time; per-seed intervals are wide ($[0.038, 0.181]$ for the worst seed at $\alpha{=}0.05$) and the ratio exceeds one in all six seed-by-budget cells. The direction repeats on the official split at counts too small to press: the 35 events lacking both pulsatile channels carry risk 0.080 at $\alpha{=}0.10$ against 0.034 to 0.038 elsewhere, and zero unsafe suppressions at $\alpha{=}0.05$.

This is not a violated guarantee but the guarantee working as specified: a marginal bound does not imply subgroup bounds. \tableref{tab:conditional} conditions on suppression, so it says that among silenced alarms a larger share was genuine in that stratum, not that a given true alarm there is likelier to be silenced; the latter needs all true alarms in the stratum as the denominator. Certifying per stratum multiplies the hypothesis count and, on strata this small, cannot pay for itself; refusing to suppress there is cheap, but it is a policy decision rather than a statistical one. It also sharpens \sectionref{sec:null}: the reliability head detects this exact stratum and still does not improve the frontier.

\section{Discussion}
\label{sec:discussion}

\paragraph{The null result.} We expected a score trained to detect incomplete or corrupted evidence to remove exactly those low-$p$ events whose low probability is untrustworthy. It does not. Three readings fit: the conditions $r$ detects are rare in VTaC; $r$ may have learned the self-generated interventions rather than the conditions under which $p$ is wrong; and the two-dimensional grid is expensive under Bonferroni. Separating them needs a naturally shifted cohort, or a target for $r$ tied to the error of $p$.

\paragraph{Limitations.} Development-arm numbers are internal; only the official-split arm is comparable to the literature, and it rests on three seeds. Record-disjoint is not patient-disjoint, and the release excludes alarms whose label was uncertain or rejected. The bound is marginal and assumes i.i.d.\ events: \sectionref{sec:conditional} shows one stratum exceeding it, and the clustered bootstrap probing that assumption is approximate, so the bound is conditional on it. No external cohort was scored, fold SDs are descriptive, and the retain threshold carries no guarantee. The grid finding is exploratory: the adaptive rule was devised after seeing the fixed grid certify poorly. Stress interventions act on the preprocessed cache. This is a retrospective research artifact: \emph{defer} is a need-for-review category, not an alert, and nothing here is validated for bedside use.

\section{Conclusion}
\label{sec:conclusion}

Risk-controlled three-way triage turns a classifier score into a policy with a finite-sample bound on the harmful endpoint, and puts a price on it: labelled alarms per budget, and candidates per grid. The second surprised us. The correction charges for every policy considered, so a grid fine enough to hold a good policy can be too fine to certify any: choosing it on the selection partition took certified folds from 1 of 15 to 8. Treat $|\Lambda|$ as a budgeted quantity. On VTaC the bound holds at 5\% in most development folds and every official seed, silencing 1.5\% of genuine alarms; reliability gating does not sharpen it.

\bibliography{ref}

\appendix

\section{Certified and Descriptive Frontiers}
\label{apd:descriptive}

\figureref{fig:frontier} is the certified frontier of \sectionref{sec:results} at full size: false-alarm suppression against realised risk on the evaluation partition for both policy families, with the budgets marked. \pOnly{} matches or exceeds \rc{} at every certified budget, which is \sectionref{sec:null} seen directly.

\begin{figure}[htbp]
\floatconts
  {fig:frontier}
  {\caption{Certified policies, development arm, against observed risk on the evaluation partition; dotted lines are the budgets, error bars SDs across seed means. Note that the two families are compared at equal \emph{budget}, not at equal observed risk, so a point lying to the left of another is more conservative rather than better. \pOnly{} matches or exceeds \rc{} at every certified budget: \sectionref{sec:null} seen directly. The point at the origin is $\alpha=0.02$, where nothing certifies and the policy suppresses nothing; its risk is undefined rather than measured at zero.}}
  {\includegraphics[width=\linewidth]{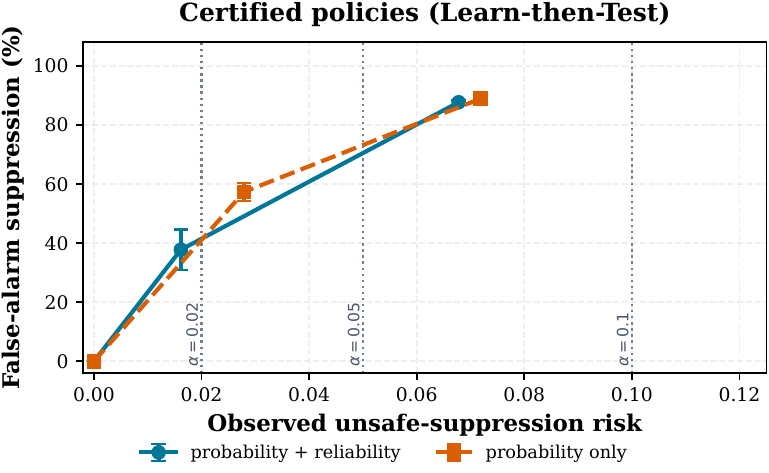}}
\end{figure}

\tableref{tab:descriptive} selects each policy on the observed risk of the calibration sample rather than on a corrected bound, so it is the frontier a practitioner would report if they skipped the correction entirely. It is included to separate two explanations of the certified frontier's shape: the model, or the selection rule. At $\alpha=0.05$ they reach 82.7\% against \rc{}'s 37.7\% and \pOnly{}'s 57.3\% in \tableref{tab:triage}, from the same frozen scores, which places the difference in the rule; at $\alpha=0.10$ the gap narrows to a few points.

\begin{table}[htbp]
\floatconts
  {tab:descriptive}
  {\caption{Policies selected on the risk \emph{point estimate} rather than the corrected bound, development arm. These carry no finite-sample guarantee and are shown only to demonstrate that the selection rule, not the model, determines the frontier: at $\alpha=0.05$ it suppresses 82.7\% against the corrected 37.7--57.3\%, though the margin narrows at $\alpha=0.10$. The uncorrected Clopper--Pearson bound on the same calibration sample nonetheless exceeds the budget in most folds. ``Cert.'' counts folds whose \emph{uncorrected} bound happened to meet the budget; no multiplicity correction was applied, so nothing here is certified.}}
  {\scriptsize
  \setlength{\tabcolsep}{2pt}
  \begin{tabular}{lcccccc}
  \toprule
  Policy & $\alpha$ & Cert. & FA supp. & Risk & Defer & CP bound \\
  \midrule
  \rc{}    & 0.02 & 0/15 & 43.1\,$\pm$\,11.2\% & 0.019 & 33.4\% & 0.159 \\
  \pOnly{} & 0.02 & 0/15 & 37.0\,$\pm$\,8.8\%  & 0.030 & 37.4\% & 0.183 \\
  \rc{}    & 0.05 & 1/15 & 82.7\,$\pm$\,0.8\%  & 0.054 & 6.1\%  & 0.059 \\
  \pOnly{} & 0.05 & 1/15 & 82.8\,$\pm$\,0.9\%  & 0.054 & 6.1\%  & 0.059 \\
  \rc{}    & 0.10 & 7/15 & 91.2\,$\pm$\,0.7\%  & 0.088 & 2.2\%  & 0.101 \\
  \pOnly{} & 0.10 & 7/15 & 91.3\,$\pm$\,0.7\%  & 0.088 & 2.1\%  & 0.101 \\
  \bottomrule
  \end{tabular}}
\end{table}

\section{Architecture and Training Details}
\label{apd:arch}

The trunk shared across slots is a two-layer strided stem (total stride 8) followed by eight depthwise-separable residual blocks, kernel 7, GroupNorm and SiLU, at width $d=512$, with block $b$ taking dilation $2^{b \bmod 6}$. Adaptive average pooling to $\lfloor L/s\rfloor$ positions forms the tokens: 375 for the 60\,s development window at $s=40$, and 312 for the 10\,s official window at $s=8$. Tokens carry a learned slot embedding, a per-slot low-rank adapter, and a sinusoidal position projection before fusion.

Modality dropout drops each available slot with probability 0.2, keeping at least one. The three auxiliary losses carry weights 0.25 (masked cross-modal reconstruction of a frozen-teacher pooled embedding for dropped slots), 0.15 (per-slot identification of a declared intervention: clipping, baseline wander, noise burst, or zero-padded shift, each applied to an available slot with probability 0.15), and 0.50 (the evidence-quality target, regressed with BCE). The model has 14.83\,M parameters in the development arm and 14.89\,M in the four-channel official arm. Training uses AdamW (learning rate $2\times10^{-4}$, weight decay 0.05, three warmup epochs, cosine decay to 1\%), effective batch 32, gradient clipping at 1.0, and at most 80 epochs with early stopping (patience 12) on checkpoint-selection AUPRC.

\section{Partition and Protocol Details}
\label{apd:protocol}

Within each development outer fold the ten record-disjoint parts are model training (5), checkpoint selection (1), policy selection (2), and policy certification (2). For fold 0 this gives 2{,}069 / 382 / 767 / 788 events with 1{,}031 held out for evaluation.

Two distinct bootstraps appear in this paper. The clustered sensitivity analysis of \sectionref{sec:certification} resamples records and takes the upper quantile of the resampled risk; it is not a second certificate. The benchmark intervals in \tableref{tab:base} are BCa \citep{efron1987better}, with the acceleration term from a jackknife over records and 20{,}000 record resamples; percentile intervals are computed alongside them and recorded in the same artifact, and no endpoint moves by more than 0.006 on AUROC or 0.7 on the Challenge Score. Every comparison drawn in \sectionref{sec:results} holds under either variant.

\paragraph{Development split.} Protocol: outer stratified group $K$-fold ($K=5$, group = waveform record) with an inner record-disjoint holdout; this is not nested cross-validation, since the inner partition is performed once. Fold-0 record counts: training 893, checkpoint selection 181, policy selection 359, policy certification 368, evaluation 459. Split file SHA-256 prefix \texttt{2afbc8ce}; assignment SHA-256 prefix \texttt{813755fb}.

\paragraph{Official split.} Protocol: official record-level partition with record-disjoint calibration cut from official train (2{,}399 training / 819 selection / 842 certification events; validation 495 events, 141 true; test 482, 137 true). The split artifact was audited event-for-event against \texttt{benchmark\_data\_split.csv} and matches the published Table~3 composition. Assignment SHA-256 prefix \texttt{485f0358}. Where the published protocol leaves a parameter unstated (notch frequency, ECG high-pass order, transition bands, choice of two ECG leads when a record carries more), our choice is recorded as a named deviation in the preprocessing configuration. The published protocol trains 10 seeds and reports the 5 best on validation; we train 3 and report all 3, so our seed spread is the more conservative.

\paragraph{Calibration and operating point.} In the development arm the base predictor has Brier score 0.107 and adaptive (equal-mass, 15-bin) ECE 0.084 \citep{nixon2019measuring}. Separately, on the official test split, sensitivity is 0.971 against specificity 0.817 at the validation-selected threshold (mean 0.115 across seeds), which reflects the Challenge Score's five-fold penalty on missed true alarms.

\paragraph{Challenge Score implementation.} Reproduces the published rule-based row exactly (67.32, F1 0.655) from its published TPR/TNR/PPV, which independently confirms both the score and the 482/137 test composition; AUROC and AUPRC agree with scikit-learn to $10^{-9}$ on 400 randomized trials with heavily tied scores.

\paragraph{Compute.} Training ran on a single NVIDIA A100 (40\,GB) with PyTorch 2.11 and TF32 matmul; the architecture avoids global attention over raw samples so that it also runs on Apple-silicon MPS for development. Best checkpoints by selection-partition AUPRC fell between epochs 7 and 31 in the development arm and 4 and 14 in the official arm.

\section{Benchmark Comparison in Full}
\label{apd:basefull}

\tableref{tab:base} carries only FCN, the strongest of the eleven systems \citet{lehman2023vtac} report in the real-time setting. \tableref{tab:basefull} restores the other ten, with the published seed spreads, so the whole comparison can be checked. FCN and CNN+CL both lie inside our record-clustered interval on every metric they report; FCN+CL's Challenge Score falls just below its lower edge, and the remaining eight fall well below.

\begin{table*}[htbp]
\floatconts
  {tab:basefull}
  {\caption{All eleven published real-time systems, the ten omitted from \tableref{tab:base} plus FCN repeated, transcribed from \citet{lehman2023vtac}, Table~4 (real-time setting), as mean\,$\pm$\,SD over their 5 best-validation seeds; the rule-based row is deterministic and carries none. Our official-split row and interval are repeated for comparison, and rows are ordered by Challenge Score. Of the eleven published systems only FCN and CNN+CL have Scores inside our interval; FCN+CL falls just below its lower edge and the remaining eight fall well below.}}
  {\small
  \setlength{\tabcolsep}{4pt}
  \begin{tabular}{lcccc}
  \toprule
  & AUROC & AUPRC & Score & F1 \\
  \midrule
  \model{} (official split) & 0.953 & 0.875 & 83.33 & 0.799 \\
  \quad 95\% CI & [0.922, 0.970] & [0.793, 0.925] & [78.5, 86.8] & [0.729, 0.852] \\
  \addlinespace
  FCN & 0.949\,$\pm$\,0.006 & -- & 80.08\,$\pm$\,2.46 & 0.805\,$\pm$\,0.016 \\
  CNN+CL & 0.943\,$\pm$\,0.005 & -- & 79.07\,$\pm$\,0.99 & 0.783\,$\pm$\,0.003 \\
  FCN+CL & 0.932\,$\pm$\,0.008 & -- & 78.41\,$\pm$\,0.87 & 0.775\,$\pm$\,0.024 \\
  CNN & 0.936\,$\pm$\,0.009 & -- & 76.17\,$\pm$\,1.20 & 0.750\,$\pm$\,0.013 \\
  SAE & 0.896\,$\pm$\,0.007 & -- & 68.77\,$\pm$\,1.11 & 0.713\,$\pm$\,0.012 \\
  Rule-based & -- & -- & 67.32 & 0.655 \\
  Transformer & 0.852\,$\pm$\,0.006 & -- & 62.73\,$\pm$\,2.78 & 0.651\,$\pm$\,0.030 \\
  Diffusion+CL & 0.685\,$\pm$\,0.017 & -- & 52.51\,$\pm$\,2.61 & 0.555\,$\pm$\,0.016 \\
  TAnoGAN & 0.657\,$\pm$\,0.012 & -- & 47.61\,$\pm$\,0.87 & 0.524\,$\pm$\,0.008 \\
  MLP & 0.706\,$\pm$\,0.008 & -- & 45.58\,$\pm$\,1.10 & 0.502\,$\pm$\,0.015 \\
  BeatGAN & 0.597\,$\pm$\,0.028 & -- & 41.02\,$\pm$\,2.71 & 0.455\,$\pm$\,0.037 \\
  \bottomrule
  \end{tabular}}
\end{table*}

\section{Reliability Head Response to Declared Interventions}
\label{apd:stress}

\figureref{fig:stress} is the full intervention sweep behind \sectionref{sec:null}, and \tableref{tab:strata} the same separation on availability as it occurs naturally in VTaC.

\begin{figure*}[htbp]
\floatconts
  {fig:stress}
  {\caption{Development arm, evaluation partitions, frozen models. (A) True-alarm probability against evidence reliability for 4{,}000 of 15{,}111 real predictions; the per-run correlation is unstable rather than consistently absent, as the inset reports. (B) Mean change in $r$ under declared interventions applied to the cached tensor, over the events each intervention can affect; error bars are SDs across 15 runs; the bar lengths are the means themselves. Gain change is a negative control, removed by z-normalization.}}
  {\includegraphics[width=0.76\textwidth]{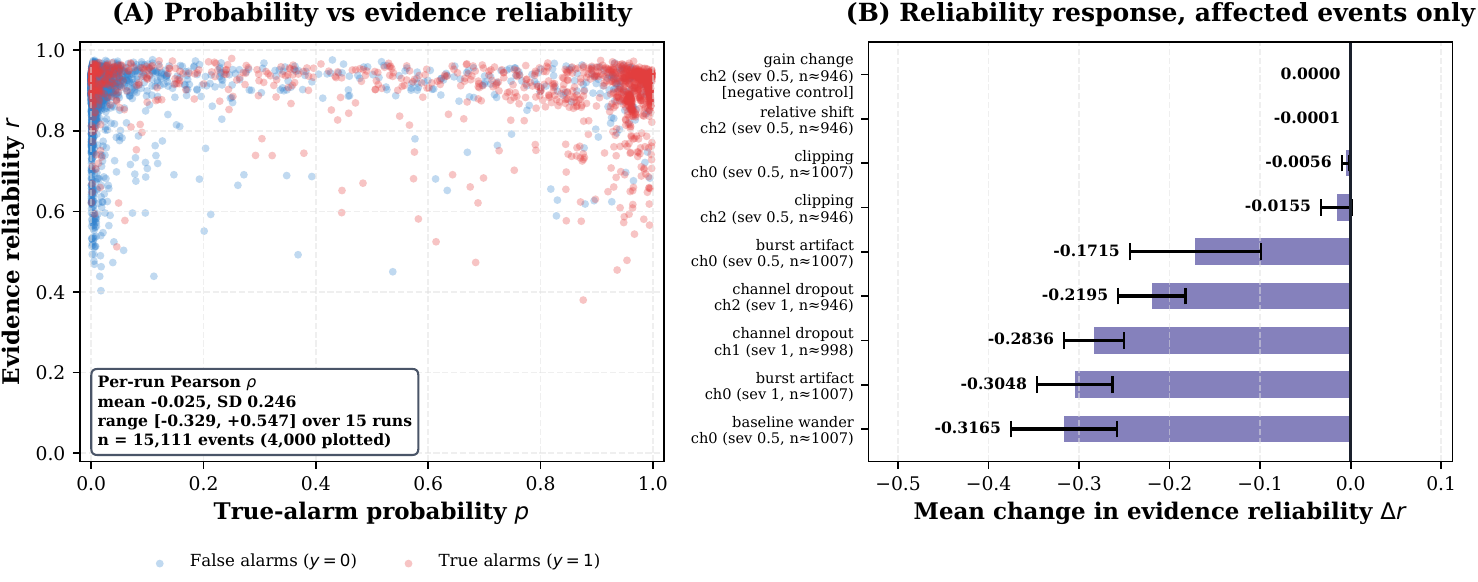}}
\end{figure*}

\begin{table}[htbp]
\floatconts
  {tab:strata}
  {\caption{Development-arm evaluation partitions by observed modality availability (mask order: ECG$_1$, ECG$_2$, pulsatile), pooled over 15 fold-runs. AUPRC/AUROC are means over runs (AUPRC $\pm$ SD); strata with fewer than 30 events in a run report no ranking metric. aECE: adaptive 15-bin ECE; $\bar r$: mean evidence reliability.}}
  {\scriptsize
  \setlength{\tabcolsep}{2pt}
  \begin{tabular}{lrrcccc}
  \toprule
  Mask & $n$ & \% & AUPRC & AUROC & aECE & $\bar r$ \\
  \midrule
  111 & 14{,}067 & 93.1 & 0.841\,$\pm$\,0.034 & 0.927 & 0.082 & 0.906 \\
  110 & 909 & 6.0 & 0.903\,$\pm$\,0.057 & 0.909 & 0.147 & 0.697 \\
  101 & 132 & 0.9 & -- & -- & -- & 0.612 \\
  100 & 3 & 0.0 & -- & -- & -- & 0.481 \\
  \bottomrule
  \end{tabular}}
\end{table}

\section{Grid Surface}
\label{apd:gridsurface}

\begin{table}[htbp]
\floatconts
  {tab:grid}
  {\caption{Held-out (\texttt{split}) certification by predeclared grid, development arm: $|\Lambda|$ candidates as $n_p$ probability by $n_r$ reliability levels. \adapt{} chooses the grid per run from the selection partition alone (\sectionref{sec:adaptive}). ``FA supp.'' is the operational average over all 15 runs, an uncertified run contributing zero, matching \tableref{tab:triage}. The released sweep also reports the admitted-only mean and the mean over selected policies regardless of certification, and the three diverge wherever runs fail: for \adapt{} at $\alpha=0.10$, whose 10 of 15 runs certify, they read 56.3\%, 84.4\% and 86.9\% respectively. Only the first is reported here, in either budget column. Nine of the 26 measured grids shown; the complete surface is recorded in the sweep artifacts (\texttt{grid\_sweep\_*.json}). None reaches $\alpha\le0.02$.}}
  {\small
  \setlength{\tabcolsep}{4pt}
  \begin{tabular}{rlrrrr}
  \toprule
  \multicolumn{2}{c}{Grid} & \multicolumn{2}{c}{$\alpha=0.05$} & \multicolumn{2}{c}{$\alpha=0.10$} \\
  \cmidrule(lr){1-2}\cmidrule(lr){3-4}\cmidrule(lr){5-6}
  $|\Lambda|$ & $n_p\!\times\!n_r$ & Cert. & FA supp. & Cert. & FA supp. \\
  \midrule
  3   & $3\!\times\!1$  & 1/15 & 4.9\%  & 5/15  & 27.4\% \\
  8   & $8\!\times\!1$  & 6/15 & 16.9\% & 10/15 & 53.3\% \\
  9   & $3\!\times\!3$  & 0/15 & 0.0\%  & 7/15  & 39.4\% \\
  30  & $30\!\times\!1$ & 7/15 & 21.4\% & 12/15 & 67.4\% \\
  36  & $12\!\times\!3$ & 3/15 & 10.0\% & 12/15 & 66.3\% \\
  59  & $59\!\times\!1$ & 6/15 & 19.9\% & 12/15 & 67.6\% \\
  60  & $12\!\times\!5$ & 2/15 & 8.0\%  & 12/15 & 66.1\% \\
  295 & $59\!\times\!5$ & 3/15 & 10.4\% & 12/15 & 63.3\% \\
  885 & $59\!\times\!15$ & 1/15 & 3.9\% & 12/15 & 62.7\% \\
  \midrule
  \multicolumn{2}{l}{\adapt{}} & \textbf{8/15} & \textbf{28.7\%} & 10/15 & 56.3\% \\
  \bottomrule
  \end{tabular}}
\end{table}

\tableref{tab:grid} shows nine of the ladder's grids. The ladder declares 28 $(n_p,n_r)$ configurations, but the sweep artifact keys them by candidate count, so $5\times3$ and $20\times3$ were overwritten by $3\times5$ and $12\times5$ and 26 configurations are reported. Each surviving row carries its own $(n_p,n_r)$ and we have checked that every one is labelled with the configuration that produced it, so nothing here is mis-attributed; but $5\times3$ and $20\times3$ are absent, and a $5\times3$ grid is a different policy family from a $3\times5$ one. Recovering them needs a re-keyed sweep, which we did not run in time for this version. The two properties below are therefore claims about the 26 configurations measured, not about all 28. The ladder is a deterministic subsampling of the same probability and reliability grids used throughout, at $n_p\in\{3,5,8,12,20,30,59\}$ crossed with $n_r\in\{1,3,5,15\}$, and the measured surface for every budget and both calibration modes is recorded in the sweep artifacts (\texttt{grid\_sweep\_*.json}), which we will release with the code. Two properties hold across all 26 configurations measured. Certified counts are not monotone in $|\Lambda|$, so a search that walks candidate counts upward and stops at the first failure will stop early. And no grid, down to three candidates, certifies $\alpha\le0.02$ on the 767-event selection partition.

\tableref{tab:gridofficial} gives the corresponding surface for the official-split arm at $\alpha=0.05$, the only budget at which grid choice changes any decision there. Both properties recur on a different model and a disjoint sample: certification is non-monotone in $|\Lambda|$, peaking at 12 candidates, and every two-dimensional grid is matched or beaten by a one-dimensional grid of comparable size.

\begin{table*}[htbp]
\floatconts
  {tab:gridofficial}
  {\caption{Official-split arm, held-out calibration at $\alpha=0.05$: certified seeds (of 3) and mean false-alarm suppression for each of the 26 distinct grid sizes on the ladder. Partitions as in \tableref{tab:triage}; evaluation is the official 482-event test set. At $\alpha=0.10$ every row certifies 3/3 and at $\alpha\le0.02$ every row certifies 0/3, so those columns are omitted.}}
  {\scriptsize
  \setlength{\tabcolsep}{3pt}
  \begin{tabular}{rlrr@{\hskip 14pt}rlrr}
  \toprule
  $|\Lambda|$ & $n_p\!\times\!n_r$ & Cert. & FA supp. & $|\Lambda|$ & $n_p\!\times\!n_r$ & Cert. & FA supp. \\
  \midrule
  3 & $3\!\times\!1$ & 0/3 & 0.0\% & 59 & $59\!\times\!1$ & 2/3 & 42.1\% \\
  5 & $5\!\times\!1$ & 1/3 & 20.9\% & 60 & $12\!\times\!5$ & 2/3 & 41.5\% \\
  8 & $8\!\times\!1$ & 1/3 & 21.9\% & 75 & $5\!\times\!15$ & 1/3 & 18.2\% \\
  9 & $3\!\times\!3$ & 0/3 & 0.0\% & 90 & $30\!\times\!3$ & 2/3 & 36.2\% \\
  12 & $12\!\times\!1$ & \textbf{3/3} & \textbf{62.7\%} & 100 & $20\!\times\!5$ & 2/3 & 42.1\% \\
  15 & $3\!\times\!5$ & 0/3 & 0.0\% & 120 & $8\!\times\!15$ & 1/3 & 19.0\% \\
  20 & $20\!\times\!1$ & 3/3 & 62.3\% & 150 & $30\!\times\!5$ & 2/3 & 34.6\% \\
  24 & $8\!\times\!3$ & 1/3 & 21.9\% & 177 & $59\!\times\!3$ & 2/3 & 41.5\% \\
  25 & $5\!\times\!5$ & 1/3 & 20.9\% & 180 & $12\!\times\!15$ & 2/3 & 41.5\% \\
  30 & $30\!\times\!1$ & 3/3 & 46.6\% & 295 & $59\!\times\!5$ & 2/3 & 40.0\% \\
  36 & $12\!\times\!3$ & 3/3 & 59.8\% & 300 & $20\!\times\!15$ & 2/3 & 42.1\% \\
  40 & $8\!\times\!5$ & 1/3 & 19.8\% & 450 & $30\!\times\!15$ & 2/3 & 35.7\% \\
  45 & $3\!\times\!15$ & 0/3 & 0.0\% & 885 & $59\!\times\!15$ & 2/3 & 41.7\% \\
  \bottomrule
  \end{tabular}}
\end{table*}

\section{Does the Grid Rule Read Held-Out Data?}
\label{apd:leak}

The adaptive rule of \sectionref{sec:adaptive} was introduced after we observed that the fixed 885-cell grid certified poorly, so its independence from held-out data should be demonstrated rather than asserted. Applying it in the pooled setting would not be legitimate: the grid would be chosen on the same sample the bound is computed on, so the family actually searched is the union of the ladder. That union is 944 distinct $(\tau_{\mathrm{sup}},\tau_{\mathrm{rel}})$ pairs, the 885-cell grid plus the 59 one-dimensional cells at $\tau_{\mathrm{rel}}{=}0$, and the smaller correction would be invalid. The pooled sweep artifact records 14/15 and 59.6\% suppression under that rule; we attach no guarantee to those figures and do not report them as results. For the held-out rule we recompute the chosen grid for all 15 fold-runs at $\alpha=0.05$ under three perturbations:

\begin{itemize}
\item permuting the \emph{evaluation} labels: every chosen grid is unchanged;
\item permuting the \emph{certification} labels: every chosen grid is unchanged;
\item permuting the \emph{selection} labels: the chosen grids change.
\end{itemize}

The first two establish that no held-out label enters the choice. The third establishes that the rule is not inert, that is, that it responds to the data it does read (7 of 15 chosen grids change under each of five selection-label permutations; 7 of the remaining 8 are runs in which no grid admits any candidate, so the rule falls back to the smallest grid regardless; the eighth chooses the 3-candidate grid on merit and the permutation leaves it there). The check (\texttt{check\_grid\_rule\_leak.py}) re-implements the ladder and the rule with numpy and scipy only, without importing the sweep code it audits, asserts that its baseline grid choices equal those recorded in the sweep artifact, and reruns in minutes from the frozen scores; its output is recorded alongside the sweep. Under pooled calibration the certification perturbation is vacuous because that partition is the selection sample, which is why the rule is reported only under \texttt{split} calibration.

\section{Calibration Volume Required}
\label{apd:calreq}

\tableref{tab:calreq} is the volume requirement behind \sectionref{sec:grid}: for each budget and candidate count, the smallest calibration sample at which some grid cell is admitted at the corrected level in \emph{every} fold-run. It is what turns an uncertifiable budget into a design parameter rather than a dead end, since it says how much more labelled data would be needed to reach it, and distinguishes budgets no sample size can reach.

\begin{table}[htbp]
\floatconts
  {tab:calreq}
  {\caption{Smallest calibration sample, in events, at which a grid cell is admitted at the corrected level in \emph{every} fold-run, from frozen evaluation predictions; parentheses give mean false-alarm suppression at that size. ``$>$8k'' marks a budget not reached up to the largest size tested; ``never'' one that no sample size reaches.}}
  {\small
  \setlength{\tabcolsep}{4pt}
  \begin{tabular}{lcccc}
  \toprule
  & \multicolumn{2}{c}{Development} & \multicolumn{2}{c}{Official} \\
  \cmidrule(lr){2-3}\cmidrule(lr){4-5}
  $\alpha$ & 885 & 59 & 885 & 59 \\
  \midrule
  0.01 & $>$8k & never$^{\ast}$ & $>$8k & $>$8k \\
  0.02 & $>$8k & $>$8k & 3k (67\%) & 2k (66\%) \\
  0.05 & 4k (67\%) & 2k (63\%) & 800 (80\%) & 500 (80\%) \\
  0.10 & 800 (82\%) & 500 (81\%) & 500 (89\%) & 500 (91\%) \\
  \bottomrule
  \multicolumn{5}{l}{\scriptsize $^{\ast}$Model-limited: risk at the minimum useful volume}\\
  \multicolumn{5}{l}{\scriptsize \phantom{$^{\ast}$}already exceeds 0.01, so no sample size suffices.}
  \end{tabular}}
\end{table}

\section{Step-Down Correction and the Confidence Level}
\label{apd:holm}

\sectionref{sec:baselines} reports that Holm's step-down procedure changes no feasibility verdict. It is not that Holm is equivalent: across the 432 run-cells it widens the admissible set in 108 and moves the selected policy in 51. The widening simply never reaches a policy that is materially better, the largest gain being 0.87 percentage points of mean false-alarm suppression. Hochberg's procedure is sharper still, but requires a positive-dependence condition we have not established for this family, so we do not rely on it.

The confidence level is the lever that does move, and it moves modestly. Lowering the family-wise confidence from 95\% to 90\% buys one to two fold-runs in five of the twenty-four arm-by-budget cells, concentrated exactly where certification is marginal: the held-out development arm at $\alpha{=}0.05$ gains 2 of 15 for \rc{} and 1 of 15 for \pOnly{}, and the held-out official arm gains its third seed. That is a trade against the strength of the claim rather than a free improvement, and it does not change the ranking of grids in \sectionref{sec:grid}.

Both comparisons are produced by \texttt{compare\_multiplicity\_corrections.py}, which rebuilds the grid, the exact bound, and the selection rule from numpy and scipy alone without importing the pipeline it audits, and which asserts that its Bonferroni feasibility verdicts reproduce those recorded in the certification artifacts: 216 cells, all matching.

\section{Selection Procedures at Every Budget}
\label{apd:procedures}

\sectionref{sec:baselines} compares Learn-then-Test against three simpler selection procedures at $\alpha=0.05$. \tableref{tab:proceduresfull} gives all three budgets. At $\alpha=0.05$ the suppression column runs in the expected order, the naive threshold suppressing most and Learn-then-Test least, but that ordering does not hold at the other budgets: at $\alpha=0.02$ split conformal suppresses more than the naive threshold, and at $\alpha=0.10$ Learn-then-Test suppresses more than split conformal. Split conformal controls a different functional in both cases, so neither reversal is a comparison of like with like. What does change monotonically with the budget is how often the other three exceed the risk they advertise.

\begin{table}[htbp]
\floatconts
  {tab:proceduresfull}
  {\caption{Four selection procedures on the same frozen scores, pooled calibration partitions, and 59-point probability grid, over 15 development fold-runs. \emph{Admitted} counts runs producing any suppression policy; \emph{In budget} counts, among those, the runs whose \emph{realised} risk on the evaluation partition met the budget. Split conformal controls a marginal quantity, so its risk column is not comparable with the others (\sectionref{sec:baselines}). A dash marks a procedure that admitted no policy, so has no risk to report. The two value columns use different denominators: ``FA supp.'' averages over all 15 fold-runs with a non-admitting run contributing zero, matching \tableref{tab:triage}, while ``Risk'' is defined only where a policy exists and so averages over admitted runs alone. That is why \tableref{tab:triage} reports 0.028 for the \pOnly{} row at $\alpha=0.05$ where this table reports 0.030: the same quantity over 15 runs rather than 14 ($0.030\times14/15$).}}
  {\scriptsize
  \setlength{\tabcolsep}{2pt}
  \begin{tabular}{llccrr}
  \toprule
  $\alpha$ & Procedure & Admitted & In budget & FA supp. & Risk \\
  \midrule
  \multirow{4}{*}{0.02}
   & Naive threshold   & 13/15 & 6/13  & 37.0\% & 0.034 \\
   & CP, uncorrected   & 6/15  & 6/6   & 12.2\% & 0.012 \\
   & Split conformal   & 15/15 & 8/15  & 45.1\% & 0.021 \\
   & \textbf{Bonferroni LTT} & 0/15 & \textbf{refuses} & 0.0\% & -- \\
  \midrule
  \multirow{4}{*}{0.05}
   & Naive threshold   & 15/15 & 5/15  & 82.8\% & 0.054 \\
   & CP, uncorrected   & 14/15 & 11/14 & 71.7\% & 0.042 \\
   & Split conformal   & 15/15 & 15/15 & 68.3\% & 0.030 \\
   & \textbf{Bonferroni LTT} & 14/15 & \textbf{14/14} & 57.3\% & 0.030 \\
  \midrule
  \multirow{4}{*}{0.10}
   & Naive threshold   & 15/15 & 10/15 & 91.3\% & 0.089 \\
   & CP, uncorrected   & 15/15 & 14/15 & 90.0\% & 0.080 \\
   & Split conformal   & 15/15 & 15/15 & 82.5\% & 0.053 \\
   & \textbf{Bonferroni LTT} & 15/15 & \textbf{15/15} & 88.9\% & 0.072 \\
  \bottomrule
  \end{tabular}}
\end{table}

\section{Realised Risk by Modality Stratum}
\label{apd:conditional}

\tableref{tab:conditional} is the per-seed detail behind \sectionref{sec:conditional}, which reports the contrast pooled over seeds. The stratum without a usable pulsatile channel is small enough that a single seed could otherwise be suspected of carrying the effect, so each is shown separately; the direction is the same in all six seed-by-budget cells.

\begin{table}[htbp]
\floatconts
  {tab:conditional}
  {\caption{Realised risk of the certified \pOnly{} policy by modality availability (mask order ECG$_1$, ECG$_2$, pulsatile), development arm. Counts pool, per seed, the outer folds in which a policy was admitted; at $\alpha{=}0.05$ seed 317 rests on four folds, one being infeasible. Strata 100 and 101 hold under 1\% of events and are omitted. \textbf{Bold}: realised risk exceeds the certified budget.}}
  {\scriptsize
  \setlength{\tabcolsep}{3pt}
  \begin{tabular}{lrrrrr}
  \toprule
  $\alpha$ & Seed & \multicolumn{2}{c}{110: no pulsatile} & \multicolumn{2}{c}{111: complete} \\
  \cmidrule(lr){3-4}\cmidrule(lr){5-6}
   &  & Unsafe/supp. & Risk & Unsafe/supp. & Risk \\
  \midrule
  \multirow{3}{*}{0.05}
   & 317  & 3/62  & 0.048 & 55/1931 & 0.028 \\
   & 911  & 7/76  & \textbf{0.092} & 69/2042 & 0.034 \\
   & 2718 & 6/75  & \textbf{0.080} & 56/2160 & 0.026 \\
  \midrule
  \multirow{3}{*}{0.10}
   & 317  & 16/150 & \textbf{0.107} & 226/3280 & 0.069 \\
   & 911  & 19/154 & \textbf{0.123} & 225/3223 & 0.070 \\
   & 2718 & 18/149 & \textbf{0.121} & 233/3274 & 0.071 \\
  \bottomrule
  \end{tabular}}
\end{table}

\section{Where the Admissible Set Sits}
\label{apd:fixedseq}

\sectionref{sec:certification} declines fixed-sequence testing on the grounds that it has no power on this family, not that it is invalid. \tableref{tab:fixedseq} is the measurement behind that claim: for each of the 18 runs across both arms, the first and last accepted index on the 59-level probability grid at $\alpha{=}0.05$, pooled calibration, Bonferroni-corrected.

\begin{table}[htbp]
\floatconts
  {tab:fixedseq}
  {\caption{Acceptance region on the one-dimensional grid at $\alpha=0.05$, pooled calibration, as indices into the 59 thresholds (0--58). ``$n$ at first'' is the suppression-set size at the first accepted threshold; ``$n$ at floor'' is the size at the smallest threshold, where a bound of 1.0 is forced by an empty suppression set rather than by risk. A run with no accepted threshold is marked \emph{none accepted} across the two index columns. Produced by \texttt{audit\_admissible\_structure.py}.}}
  {\scriptsize
  \setlength{\tabcolsep}{2pt}
  \begin{tabular}{lrrrr}
  \toprule
  Run & First & Last & $n$ at first & $n$ at floor \\
  \midrule
  dev outer0 seed2718 & 5  & 19 & 194 & 0 \\
  dev outer0 seed317  & 5  & 19 & 212 & 0 \\
  dev outer0 seed911  & 15 & 22 & 168 & 0 \\
  dev outer1 seed2718 & 20 & 22 & 547 & 0 \\
  dev outer1 seed317  & 14 & 20 & 143 & 0 \\
  dev outer1 seed911  & 20 & 20 & 303 & 0 \\
  dev outer2 seed2718 & 15 & 20 & 338 & 0 \\
  dev outer2 seed317  & 4  & 5  & 609 & 0 \\
  dev outer2 seed911  & 1  & 3  & 233 & 0 \\
  dev outer3 seed2718 & 4  & 4  & 535 & 0 \\
  dev outer3 seed317  & 0  & 0  & 915 & 915 \\
  dev outer3 seed911  & 4  & 6  & 462 & 0 \\
  dev outer4 seed2718 & 1  & 7  & 371 & 32 \\
  dev outer4 seed317  & \multicolumn{2}{c}{none accepted} & -- & 980 \\
  dev outer4 seed911  & 5  & 21 & 268 & 0 \\
  official seed2718   & 4  & 20 & 213 & 0 \\
  official seed317    & 20 & 23 & 578 & 0 \\
  official seed911    & 20 & 26 & 381 & 0 \\
  \bottomrule
  \end{tabular}}
\end{table}

These regions are computed at the corrected level $\delta/59$, so they bound what a \emph{corrected} walk can reach; a fixed-sequence procedure would test each hypothesis it reaches at $\delta$, and a threshold rejected here could be accepted there. Only the low-end result transfers, because it does not depend on the level at all: 15 of the 18 runs suppress \emph{zero} events at the smallest threshold, which forces a Clopper--Pearson upper bound of 1.0 by construction, before risk is consulted and whatever the level. An ascending walk therefore stops at its first step in those 15 runs under any correction. The descending order, and the three non-empty ascending endpoints, would need direct $\delta$-level tests we did not run.

At the corrected level the accepted set is contiguous in all 17 runs where it is non-empty, it begins at the smallest threshold in only 1 of those 17, and it reaches the largest in none. An earlier implementation of this work under the ascending fixed-sequence rule accordingly certified 0 of 45 fold--budget combinations. The one run that begins at index 0 is also the one whose smallest threshold already suppresses 915 events, which is the exception that states the rule: an ascending walk can only take a second step when the grid floor is coarse enough to have accumulated a usable sample. Declaring a coarser grid is exactly the remedy \sectionref{sec:grid} arrives at from the other direction, and the two observations are the same fact seen twice.

\section{Correction Record}
\label{apd:corrections}

A pre-submission audit of an earlier draft found: (i) three of four figures contained values not traceable to any stored artifact, including a scatter generated from a Beta distribution and an availability-stratum ordering that was inverted relative to the real values; (ii) a ``safety guarantee'' sentence hard-coded in a report template that the run log contradicted; (iii) selection on the observed risk followed by certification on the upper bound, which does not control risk; (iv) a model trained on 20\% of the data because of the inner partition allocation; (v) the certification script re-deriving the split at runtime rather than loading the committed artifact; (vi) a deferral rate that was a grid-ordering artifact; (vii) a comparator with a different grid and no deferral mechanism; (viii) a fixed-sequence Learn-then-Test rule that admitted nothing (\sectionref{sec:certification}); and (ix) an ``external validation'' script that scored no external data. All are corrected in the code we will release. Reporting code now reads only from the generated input artifact and raises on a missing value rather than substituting a placeholder; withdrawn artifacts are retained in a quarantined directory and are not cited here. A later audit of the compiled PDF found a tenth: a corrected table body written into the wrong float, so one appendix table printed grid-sweep rows under a policy caption while the grid table kept superseded values. Both are regenerated here, and the verification script now parses the rendered table bodies back out of the source rather than only checking that each artifact matches a literal. Known and unaddressed: the engineered-feature comparator receives unstandardized descriptors, and absolute-amplitude features could act as a device shortcut; both belong to comparators not reported in this paper.

\end{document}